\pdfoutput=1
\documentclass[11pt]{article}

\usepackage[final]{coling}
\usepackage{times}
\usepackage{latexsym}
\usepackage[T1]{fontenc}
\usepackage[utf8]{inputenc}
\usepackage{microtype}

\usepackage{graphicx}
\usepackage{amsmath}
\usepackage{amssymb}
\usepackage{tikz}
\usepackage{pgfplots}
\usepackage{booktabs}
\usepackage{multirow}
\usetikzlibrary{positioning,calc}%
\pgfplotsset{compat=1.18}
\usepackage{layout}
\newcommand{\reds}[1]{\textcolor{black}{#1}}
\newcommand{\cyans}[1]{\textcolor{black}{#1}}

\usepackage{hyperref}
\usepackage{inconsolata}
\usepackage{graphicx}

\DeclareUnicodeCharacter{2796}{\textendash}

\title{Early Exit Is a Natural Capability in Transformer-based Models: An Empirical Study on Early Exit without Joint Optimization}
\author{\ \ \ \ \ \ \ \ \ \ \ \ \ \ \ \ Weiqiao Shan\textsuperscript{1}\footnotemark[1], Long Meng\textsuperscript{1}\footnotemark[1], Tong Zheng\textsuperscript{1}, Yingfeng Luo\textsuperscript{1}, Bei Li\textsuperscript{3},\\ \textbf{junxin Wang}\textsuperscript{4}, \textbf{Tong Xiao}\textsuperscript{1,2}, \textbf{Jingbo Zhu}\textsuperscript{1,2}\\
\textsuperscript{1}School of Computer Science and Engineering, Northeastern University, Shenyang, China\\
\textsuperscript{2}NiuTrans Research, Shenyang, China\\
\textsuperscript{3}Meituan, Beijing, China\\
\textsuperscript{4}Dalian Jiaotong University, Shenyang, China}

\begin{document}
\maketitle

\footnotetext[1]{Equal contribution.}

\begin{abstract}

Large language models (LLMs) exhibit exceptional performance across various downstream tasks. However, they encounter limitations due to slow inference speeds stemming from their extensive parameters. The early exit (EE) is an approach that aims to accelerate auto-regressive decoding. EE generates outputs from intermediate layers instead of using the whole model, which offers a promising solution to this challenge. However, additional output layers and joint optimization used in conventional EE hinder the application of EE in LLMs. 

In this paper, we explore the possibility of LLMs EE without additional output layers and joint optimization. Our findings indicate that EE is a natural capability within transformer-based models. While joint optimization does not give model EE capability, it must be employed to address challenges by improving the accuracy of locating the optimal EE layer through gating functions. Additionally, our study reveals patterns in EE behavior from a sub-word perspective based on the LLaMA model and the potential possibility for EE based on sub-layers.

\end{abstract}

\section{Introduction}

Recently, Large language models (LLMs) have witnessed widespread adoption in Natural Language Processing (NLP)~\cite{brown2020language,openai2023gpt4, touvron2023llama}. Owing to their extensive parameter count, LLMs demonstrate strong capabilities in handling natural language understanding tasks and have achieved notable success in natural language generation tasks, such as machine translation (MT)~\cite{xu2023paradigm}. However, their extensive parameter also leads to unaffordable computation costs and latency during generation.

\begin{figure}[tp]
    \centering
    \includegraphics[width=0.48\textwidth]{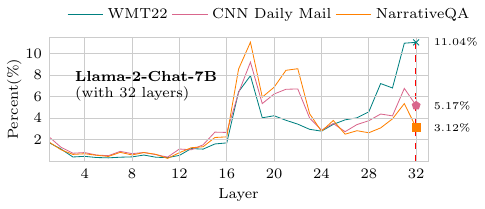}
    \caption{Percentage of tokens at each layer across the entire test set where a token first matches the final output during inference.}
    \label{fig:max-early-exit-in-diff-datasets}
\vspace{-1em}
\end{figure}

\begin{figure*}[tp]
    \centering
    \includegraphics[width=1.0\textwidth]{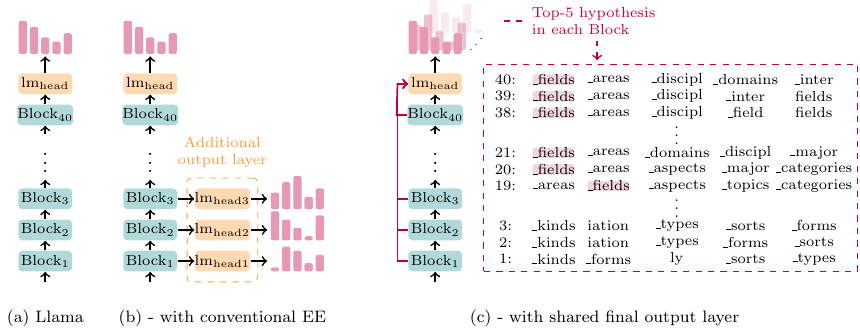}
    \caption{Example of Llama-2-Chat-13B when doing translation task on WMT22-ZH2EN test set, we found that the Top-1 hypothesis is same with the final Top-1 output far away from the final layer($\text{Block}_{40}$).}
    \label{fig:saturation-event-in-llm}
\end{figure*}

To address the substantial computational costs, inference can be accelerated and made more efficient through early exit, which leverages only a subset of layers in a model~\cite{teerapittayanon2016branchynet}. Unlike static acceleration methods like pruning, this dynamic decoding strategy enables the model to adjust its computational requirements based on the difficulty of the input~\cite{elbayad2019depth, liu2021faster}. In conventional early exit models, multiple additional output layers are introduced, as shown in Fig.~\ref{fig:saturation-event-in-llm}(a) and (b), and a careful finetuning stage called joint optimization is used to ensure optimal performance~\cite{elbayad2019depth, schuster2021consistent, schuster2022confident}. 

Notably, this conventional early exit framework is insufficient for LLMs attributed to a key limitation: Each early exit layer needs an additional output layer. With more early exit layers, a higher speed-up can be obtained, but more additional output layers should be added. This additional output layer not only significantly increases the model parameters~\cite{kavehzadeh2023sorted, bae2023fast}, but also requires joint optimization, which is computationally expensive and may degrade the performance of model benefit from the pre-trained~\cite{xin2020deebert,xin2021berxit}.

Recent studies attempt to avoid introducing additional parameters by either utilizing only the final output layer~\cite{del2023skipdecode, elhoushi2024layer} or significantly reducing the number of layers that can exit~\cite{kavehzadeh2024sorted}. While these approaches reduce reliance on the additional output layers, they do not fully address the core dependence on joint optimization. This challenge has prompted our investigation into the necessity of joint optimization, particularly in the context of LLMs. Specifically, we aim to explore whether a model can generate tokens at shallower layers that are consistent with those from the final layer, without additional modules and fine-tuning stage, and whether existing early exit approaches can exploit this capability. Notably, in our preliminary experiments with the vanilla LLaMA model~\cite{touvron2023llama2} across various generation tasks, we observed that no more than 11.08\% tokens require processing by the final layer, as shown in Fig.~\ref{fig:max-early-exit-in-diff-datasets}.

We extended this experiment to various transformer-based models, including both encoder-decoder and encoder-only architectures, and yielded the same conclusion: early exit is an intrinsic capability of transformer-based models and does not depend on joint optimization. However, we observed that transformer-based models have early exit capability naturally, this capability is not derived from joint optimization. The joint optimization only improves the effectiveness of existing gating functions by increasing the similarity of the distributions between neighboring layers, despite its negative impact on the performance of the full model.

Additionally, we find that implementing early exit in the sequence-level scenario is straightforward, whereas it presents significant challenges in a token-level scenario due to the absence of key and value information from previous tokens required for attention operations. While previous approaches that copy key-value pairs (KV) help mitigate this issue, they introduce error propagation during inference, making it particularly challenging for longer sentence generation. It's worth noting that, the model with joint optimization appears to break the intrinsic connection between the KV from neighboring layers, performing worse than the model without joint optimization when KV copy operations are involved. This observation motivates us to explore the underlying properties of early exit capability further, aiming to develop gating functions that are independent of joint optimization.

\section{Background}

\subsection{Large Language Models}
Large language model constructed based on Transformer decoder architecture with $N$ blocks~\cite{vaswani2017attention}, each block consists of three sub-layers multi-head self-attention (attn), feedforward network (ffn) and layer normalization, which commonly based on the pre-norm architecture. The final hidden state of model can be represented by $h_{t}^{\ell}=\operatorname{block}_{\ell}(h_{\leq t}^{\ell-1})$, in which:
\begin{eqnarray}
& & {h_{\text{attn}}}^{\ell}_t = \text{attn}({h}^{\ell-1}_t) + {h}^{\ell-1}_t  \\ \nonumber
& & h_{t}^{\ell} = \text{ffn}({h_{\text{attn}}}^{\ell}_t) + {h_{\text{attn}}}^{\ell}_t, \; \ell \in N
\end{eqnarray}

Where ${h}^{\ell}_t$ is the output of block $\ell$ for token $t$, for brevity, we omitted all layer normalization in the model. Based on the ${h}^{N}_t$, a distribution from an output layer ($\text{lm}_\text{head}$) can be parameterized by ${W}_{N}$:
\begin{equation}
p(y^{N}_{t}) = \operatorname{softmax}({W}_{N} {h}^{N}_t)
\label{eq:func-prediction-by-lm-head}
\end{equation}

\subsection{Early Exit}\label{sec:early-exit-back}

\paragraph{Architecture} Early exit models often consist of a backbone network with additional output layers and a gating function, where the backbone network is commonly used base model such as Transformer and LLaMA. The additional output layer provides loss information during training and provides the ability that output at the current layer during inference. The gating function is used to guide the network when to exit.

\paragraph{Joint optimization} Based on a pre-trained or randomly initialized backbone model, conventional early exit involves attaching multiple output layers to the backbone model and optimizing together, called joint optimization, represented as $\mathcal{L}=\sum_{l=1}^N Loss_l$. Some methods will optimize the gating network individually after joint optimization.

\paragraph{Inference} To obtain the ideal exit layer for each token during the inference stage, most works decide the exit layer by a gating function that compares a value derived from the output layer with a threshold $\tau$, which is sensitive to the distribution output from each layer and an additional fine-tuning stage.
\begin{itemize}
\item {Confidence-based~\cite{liao2021global}}. Exit at layer $i$ when the max output probability exceeds the threshold, $\max{p(y^{i}_{t})} > \tau$.

\item {Entropy-based~\cite{xin2020deebert}}. Exit at layer $i$ when the entropy of the output probability distribution is lower than the threshold, $\text{entropy}(p(y^{i}_{t})) < \tau$.

\item {Patience-based~\cite{zhou2020bert}}. Exit at layer $i$ when the output token in continous $\tau$ layer is same ($\arg\max_{tok}p(y^{i}_{t}) = \ldots = \arg\max_{tok}p(y^{i-\tau}_{t})$)\footnote{$tok$ with max probability in each layer distribution $p(y^{\cdot}_{t})$.}.
\end{itemize}
To explore the upper bound of the early exit layer in the model, we recognize the optimal early exit layer, which can exit at the earliest layer while having the same output as the final layer by an ideal gating function. 
\begin{itemize}
\item {Ideal}. Exit at layer $i$ when the predicted token is the same as the final layer, $\arg\max_{tok}p(y^{i}_{t}) = \arg\max_{tok}p(y^{N}_{t})$.
\end{itemize}

\paragraph{Copying the KV cache} In the sequence-level early exit approach, all tokens in one sequence exit at the same layer,  making early exit implementation straightforward. However, in the token-level early exit approach, each token may exit at a different layer, presenting a significant challenge for auto-regressive decoding. When compute the output ${h}^{i}_t$ for $t$-th token at $i$-th layer, self-attention operation requires access to all $\mathbf{K}^{i}_{<t}$ and $\mathbf{V}^{i}_{<t}$. Unfortunately, a likely scenario arises where some previous tokens may exit at an earlier layer, implying that obtaining $\mathbf{K}^{i}_{<t}$ can only be approximated by $\mathbf{K}^{<i}_{<t}$ or $\operatorname{ATTN_i}({h}^{<i}_{<t})$. This scenario poses a challenge as it limits the availability of information for the auto-regressive decoding process.

\section{Experimental Setup}

We experiment on three types of transformer-based models, including the BERT and RoBERTa (encoder-only) model, the transformer-based (encoder-decoder) model, and the Llama2 (decoder-only) model. All experiments were conducted using the Transformers\footnote{\url{https://github.com/huggingface/transformers}} and Fairseq\footnote{\url{https://github.com/facebookresearch/fairseq}} toolkit based on the GeForce RTX 3090 * 8, and A800 80GB * 4 only for Llama fine-tuning. 

\paragraph{Dataset} We select GLUE benchmark for encoder-only model and use toolkit \textit{glue\_compute\_metrics} for Acc/F1. For the transformer-base model, we do experiments on WMT14 EN2DE translation tasks and compute BLEU and COMET following~\cite{wang2019learning, zheng2023partialformer, li2024predictor}. For the Llama model, we experiment on three generation test sets, including WMT22, CNN Daily Mail, and NarrativeQA, and we select COMET~\cite{rei2020comet} with \texttt{Unbabel/wmt22-comet-da} and ROUGE-L~\cite{lin2004rouge} as the metrics separately.

\paragraph{Model} We explore the early exit capability on the pre-trained backbone model and the pre-trained model with joint optimization separately. For training the backbone model we follow the work which is widely accepted\footnote{\url{https://github.com/huggingface/transformers/tree/v4.37.2\\/examples/pytorch/text-classification}}~\cite{wang2019learning, elbayad2019depth}. For the joint optimization stage, we follow~\cite {teerapittayanon2016branchynet, alpaca} without additional output layers, which means we obtain output from each layer all based on the final output layer. During the decoding stage, the prompt we used in LLMs is listed in \ref{sec:template-for-prompt} and select Top-$k$ sampling for all generation tasks.

\section{Overview}

\subsection{Early Exit Is A Natural Capability In LLMs Without Joint Optimization}\label{sec:Early-Exit-in-LLMs}

Conventional early exit incorporates multiple additional output layers, which result in redundant parameters and require an additional training stage. On one hand, this approach significantly increases the parameter count due to the larger hidden dimension and extensive vocabulary size. On the other hand, joint optimization may cause LLMs to forget the knowledge acquired during the pre-training stage, further hindering the practical application of early exit in LLMs. This challenge has motivated our investigation into whether a natural early exit capability exists in decoder-only models that leverage shared final output layers without joint optimization.

\paragraph{Experiment} We evaluate the optimal early exit layer based on the \texttt{Llama-2-7b-chat}\footnote{\url{https://huggingface.co/meta-llama/Llama-2-7b-chat}} and \texttt{Llama-2-13b-chat}\footnote{\url{https://huggingface.co/meta-llama/Llama-2-13b-chat}} models across ten language pairs from the WMT22 machine translation benchmark, the NarrativeQA dataset for question answering, and the CNN/DailyMail dataset for summarization tasks. To explore the upper bound of natural early exit capacity in the Llama model under the ideal situation, we employ Top-1 sampling and utilize the full model for every token. Ensuring that no early exit occurs, thus avoiding the need for copying the KV cache and preventing the impact of error propagation during auto-regressive decoding.

\paragraph{Results} Remarkably, our experiments demonstrate that the output of intermediate layers starts to match the oracle final output before reaching the final layer at the last ten layers across various tasks\footnote{We have also experimented on the commonly used datasets for LLMs, and lead consistent conclusions, in detail at \ref{sec:commonly-used-dataset}}, as shown in Tab.~\ref{tab:optimal-early-exit-layer-in-llama2}. This is very similar to the saturation phenomenon mentioned in previous work~\cite{geva2022transformer}. However, we observe a stronger consistency with the top-ranked candidates at each layer, where these tokens maintain a stable relative order across multiple layers, as illustrated in Fig.~\ref{fig:saturation-event-in-llm} (c). This observation underscores early exit capability in LLMs, independent of additional output layers and joint optimization, due to the natural redundancy within the model.

\begin{table}[htp]
\small
\centering
\begin{tabular}{l|c|c|c|c}
\toprule
\multirow{2}{*}{Model} & \multicolumn{3}{c|}{Early exit Layer} & WMT22 \\ \cline{2-5}
& Max & Avg & Perc & COMET-22 \\
\midrule
7B & \textbf{32} & \textbf{23.49} & 88.23\% & 79.68\textbackslash79.71 \\
13B & \textbf{40} & \textbf{25.5} & 92.43\% & 80.65\textbackslash80.68 \\
\midrule
\multicolumn{5}{c}{Rough-1\textbackslash-2\textbackslash-L on CNN{\_}DM} \\
\midrule
7B & \textbf{32} & \textbf{21.11} & 94.83\% & 20.39\textbackslash8.07\textbackslash19.76 \\
13B & \textbf{40} & \textbf{24.29} & 64.45\% & 20.45\textbackslash8.12\textbackslash19.82 \\
\midrule
\multicolumn{5}{c}{Rough-1\textbackslash-2\textbackslash-L on NarrativeQA} \\
\midrule
7B & \textbf{32} & \textbf{20.67} & 96.88\% & 25.16\textbackslash11.05\textbackslash23.84 \\
13B & \textbf{40} & \textbf{23.47} & 71.72\% & 24.83\textbackslash11.05\textbackslash23.62 \\
\bottomrule
\end{tabular}
\caption{We report the average optimal early exit layer in multiple directions in WMT22 (detailed in \ref{sec:saturation-statistical-information}). The \texttt{Avg} represents the average optimal early exit layer, and the \texttt{Perc} is the percentage of tokens that can early exit in all tokens.}

\label{tab:optimal-early-exit-layer-in-llama2}
\end{table}

\begin{table}[htp]
\small
\centering
\begin{tabular}{l|c|c|c|c|c}
\toprule
Task & layer & avg & perc & BLEU & COMET\\
\midrule
WMT & 6-\textbf{6} & \textbf{4.92} & 57.25\% & 26.75 & 83.86 \\
IWSLT & 6-\textbf{6} & \textbf{4.18} & 80.01\% & 32.18 & 70.53 \\
\bottomrule
\end{tabular}
\caption{The optimal early exit layer for a decoder in Transformer-Base model on WMT14 EN2DE and IWSLT14 DE2EN.}
\label{tab:optimal-early-exit-layer-in-transformer}
\end{table}

\subsection{Early Exit Capability Is Universal In Various Transformer-based Models}

Based on previous results, a natural question is whether the phenomenon is universal or only occurs in decoder-only models like LLMs.

\paragraph{Experiment} We expand our experiment on multiple models and tasks in which the models have entirely different structures. Including the Transformer-base model, BERT, and RoBERTa with the WMT14-DE2EN dataset and GLUE benchmark.

\begin{table*}[htp]
\small
\centering
\begin{tabular}{l|c|l|l|l|l|l|l|l|l}
\toprule
Model & layer & Metric & CoLA & MRPC & QNLI & QQP & RTE & MNLI & SST-2 \\ 
\midrule
\multirow{3}{*}{RoBERTa} & \multirow{3}{*}{\textbf{12}} & F1/Acc & 56.24(Mcc) & 91.28/88.23 & 93.09 & 91.32/88.50 & 72.56 & 87.76 & 94.26 \\
&  & Avg & \textbf{3.17} & \textbf{1.32} & \textbf{5.44} & \textbf{5.5} & \textbf{8} & \textbf{7.01} & \textbf{3.83} \\
&  & Perc & 81.11\% & 100\% & 92.97\% & 99.82\% & 77.26\% & 98.68\% & 100\% \\
\midrule
\multirow{3}{*}{BERT} & \multirow{3}{*}{\textbf{12}} & F1/Acc & 56.49(Mcc) & 87.32/82.60 & 91.61 & 91.12/88.07 & 70.03 & 84.87 & 92.88\\
&  & Avg & \textbf{3.11} & \textbf{2.88} & \textbf{2.87} & \textbf{2.04} & \textbf{4.34} & \textbf{6.46} & \textbf{4.67} \\
&  & Perc & 93.1\% & 93.68\% & 99.65\% & 100\% & 96.39\% & 93.4\% & 99.77\% \\
\bottomrule
\end{tabular}
 \caption{Result in RoBERTa-base and BERT-base-uncased model.}
\label{tab:optimal-early-exit-layer-in-bert}
\end{table*}

\begin{table*}[htp]
\small
\centering
\begin{tabular}{l|c|l|l|l|l|l|l|l|l}
\toprule
Model & joint & Metric & CoLA & MRPC & QNLI & QQP & RTE & MNLI & SST-2 \\ 
\midrule
\multirow{3}{*}{RoBERTa} & without & Spd-up & 2.52$\times$ & 3.52$\times$ & 1.80$\times$ & 1.76$\times$ & 2.07$\times$ & 1.48$\times$ & 2.43$\times$ \\
& with & Spd-up & \cyans{2.96}$\times$ & \cyans{2.59}$\times$ & \cyans{2.85}$\times$ & \cyans{3.08}$\times$ & \cyans{2.93}$\times$ & \cyans{2.59}$\times$ & \cyans{2.94}$\times$ \\
& with & F1/Acc & \reds{54.29}(Mcc) & \reds{90.23}/\reds{86.78} & \reds{92.53} & \reds{88.14}/\cyans{91.11} & \reds{67.87} & \reds{87.59} & \reds{94.04} \\
\midrule
\multirow{3}{*}{BERT} & without & Spd-up & 2.71$\times$ & 2.77$\times$ & 2.69$\times$ & 3.18$\times$ & 2.69$\times$ & 1.63$\times$ & 2.17$\times$ \\
& with & Spd-up & \cyans{3.07}$\times$ & \cyans{3.22}$\times$ & \cyans{3.30}$\times$ & \cyans{3.52}$\times$ & \cyans{2.87}$\times$ & \cyans{3.00}$\times$ & \cyans{3.59}$\times$ \\
& with & F1/Acc & \reds{56.24}(Mcc) & \reds{85.08}/\reds{79.01} & \reds{90.55} & \reds{87.56}/\cyans{90.83} & \reds{62.09} & \reds{83.78} & \reds{91.86} \\
\bottomrule
\end{tabular}
\caption{Real speed-up of RoBERTa and BERT model with and without joint (joint optimization) under the ideal gating functions, We find a significant decrease in the performance of the model with joint optimization, compared to the baseline results in Tab.~\ref{tab:optimal-early-exit-layer-in-bert}.}
\label{tab:use-saturation-in-bert-model}
\end{table*}

\paragraph{Results} We found that the early exit capability exists across different models and tasks. In the encoder-only model, early exit seems to be very common due to the glue benchmark tending to be binary classification tasks, under some classification tasks, nearly all tokens only require less than one-sixth of layers to get the same result as using the full model, like RoBERTa on MRPC or BERT on QQP, as shown in Tab.~\ref{tab:optimal-early-exit-layer-in-bert}. Whereas in the encoder-decoder model, the fewer layers lead to less redundancy in the model, so that on more difficult tasks there is less early exit space for the Transformer model, as shown in Tab.~\ref{tab:optimal-early-exit-layer-in-transformer}.

\section{Can Early Exit Capability Be Utilized Directly?}

\subsection{Early Exit Requires Joint Optimization Even In The Simplest Scenarios}\label{sec:Can-Early-Exit-Capability-Be-Used-Directly}

\paragraph{Motivating} Building on the early exit capability, two important questions arise: 1) Can we leverage this capability directly to enhance decoding efficiency? and 2) Can the gating function used in previous works accurately identify the earliest exit layer? To address these questions, We first conduct experiments on the GLUE benchmark using BERT and RoBERTa models in the sequence-level early exit scenario. Exploiting the early exit capability in this context is particularly straightforward, the sequence-level early exit scenario does not involve auto-regressive decoding and thus bypasses the need for copying the KV cache. 

\paragraph{Experiment} For a fair comparison, the pre-trained backbone model was obtained following previous works\footnote{\url{https://github.com/huggingface/transformers/tree/v4.37.2\\/examples/pytorch/text-classification}} and the model after joint optimization without additional output layer follow the BranchyNet~\cite{teerapittayanon2016branchynet}. In terms of the gating function, we compare confidence-base, entropy-based, and patience-base gating functions mentioned above, which these gating functions are not limited by the model structure.

\begin{figure}[tp]
    \centering
    \includegraphics[width=0.45\textwidth]{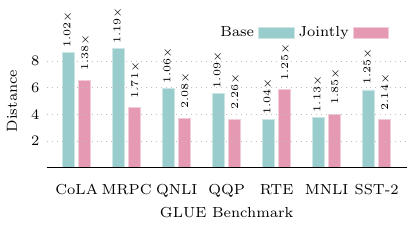}
    \caption{The average distance and speed-up between the optimal early exit layer and the exit layer from the three gating functions in RoBERTa model(more details in~\ref{sec:appendix-gating-function directly-on-bert-like-model}). We constrain the performance of early exit not less than 98\% original model to obtain the optimal threshold for gating functions.}
    \label{fig:direct-exit-by-previous-function}
\end{figure}

\paragraph{Results} We find that identifying the optimal early exit layer without joint optimization is a formidable challenge. Based on the above three gating functions at the optimal thresholds\footnote{We find the optimal thresholds through grid search.}, we compared the distance between the early exit layer selected by the gating function and the optimal exit layer. Our results indicate that, in the model without joint optimization, the distance is greater and seems unable to bring acceleration actually, as shown in Fig.~\ref{fig:direct-exit-by-previous-function}. This suggests that joint optimization assists the gating network in more accurately identifying layers suitable for early exit, while still ensuring the decoding results. 

Additionally, we measured the actual speed-up achieved through early exit under the ideal gating function, and found that the models with joint optimization also exhibited a higher speed-up, as shown in Tab.~\ref{tab:use-saturation-in-bert-model}. This suggests that joint optimization not only improves the accuracy of the gating network but also expands the space where early exit can occur, although this comes at the cost of reduced overall model performance. Notably, we obtained similar results with Transformer models (detailed in~\ref{sec:appendix-gating-function directly-on-bert-like-model}), but in the LLaMa model, a slight speed-up was achieved even without joint optimization (detailed in~\ref{sec:appendix-gating-function directly-on-transformer-and-llm}). This is likely due to the inherent redundancy between each layer in LLMs.

\subsection{Joint Optimization Helps Gating Functions by Boosting Output Distribution Similarity}\label{sec:Joint-Helps-Gating-by-Boosting-Output-Distribution-Similarity}

\begin{figure}[t]
    \centering
    \includegraphics[width=0.45\textwidth]{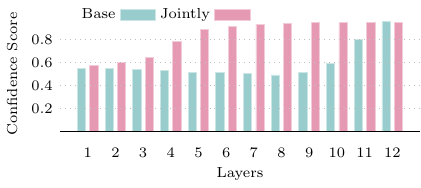}
    \caption{The average confidence score between each layer overall sample in the GLUE benchmark based on the RoBERTa model.}
    \label{fig:confidence-score-between-each-layer}
\end{figure}

Based on previous descriptions, existing gating functions rely on the output distribution at each layer. To understand how joint optimization enhances the performance of the gating function, we examined the maximum confidence scores of the output distribution in the models with and without joint optimization, as shown in Fig.~\ref{fig:confidence-score-between-each-layer}. In the models without joint optimization, the maximum confidence score is quite similar in the shallow layers and varies more significantly in the deeper layers. In contrast, models with joint optimization exhibit greater variation in confidence scores in the shallow layers, causing the gating function to favor earlier exit points. Additionally, the maximum confidence score of the model with joint optimization increases slower more gradually as the layers increase, this reduces the sensitivity of the model to the fixed threshold and further improves the effectiveness of such gating functions.

\section{Can Early Exit Capability Be Utilized With Joint Optimization In Token-level Scenario?}

\subsection{Copy KV Causes Error Propagation When Processing Long Sequence}\label{sec:Saturation-Events-for-Early-Exit-in-LLMs}

\paragraph{Experiment} To explore whether early exit can accelerate LLMs decoding in real scenarios, we perform token-level early exit experiments with joint optimization to enhance the performance of the gating function. We fine-tuning the LLaMA model with joint optimization following~\cite{alpaca}. As described in Sec.~\ref{sec:early-exit-back}, token-level early exit requires copying $\mathbf{h}^{\ell-1}_t$ and recomputing the K and V for the self-attention operation~\cite{elbayad2019depth,schuster2022confident}.

\begin{table*}[ht]
\centering
\footnotesize
\begin{tabular}{lp{4.7cm}p{9.0cm}}
\toprule
 & \textbf{Hypothesis} & \textbf{Optimal early exit layer} \\
\midrule
Full Model & \multicolumn{2}{l}{The item cost less than 20 euros.</s>} \\
Recompute & The item cost less than 20 euros.</s> & \texttt{[31, 31, 31, 31, 4, 7, 6, 6, 16, 1, 11, 7]} \\ 
\midrule
Full Model & \multicolumn{2}{l}{If something were to happen like a fire, break-in, earthquake, or alien invasion, it would be too much} \\  
 & \multicolumn{2}{l}{responsibility for K1 to take care of K2.</s>} \\
Recompute & If something were to happen like a fire, a (catastrophic) alien invasion, (a)n (alien) invasion, (a)nd (al (alien), (al (al), (al (al), (al (al), (al (al), (al $\ldots$ & \texttt{[31, 31, 31, 31, 5, 16, 7, 10, 7, 7, 31, 31, 8, 2, 2, 8, 31, 0, 7, 7, 15, 23, 5, 30, 30, 30, 13, 7, 15, 7, 6, 6, 4, 24, 7, 8, 31, 12, 15, 29, 30, 12, 31, 9, 31, 6, 12, 17, 9, 7, 6, 12, 7, 9, 7, 6, 11, 7, 9, 7, 6, 11, 7, 9, 7, 6, $\ldots$]} \\ 
\bottomrule
\end{tabular}
\caption{Token-level early exit result and exit layer of Llama-2-Chat-7B with joint optimization on WMT22-DE2EN test set.}
\label{table:kv-example-short}
\end{table*}

\paragraph{Results} Our observations indicate that early exit based on the optimal early exit layer is effective for shorter sentences. However, when applied to long sequences, it tends to generate repeat tokens and fails to terminate the sequence, as shown in Tab.~\ref{table:kv-example-short}. In particular, we find that recomputing KV in the model with joint optimization performs worse than the model without joint optimization, which illustrates another drawback of joint optimization that it destroys the intrinsic connection between KV pairs across neighboring layers (detailed shown in Tab.~\ref{table:kv-example-in-long-sentence}).

\subsection{Exploring Potentially Early Exit Functions}

\begin{figure}[htp]
    \centering
    \includegraphics[width=0.45\textwidth]{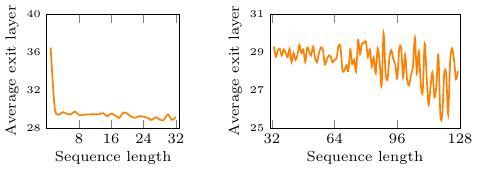}
    \caption{The relationship between the average early exit layer and sequence length based on the Llama-2-Chat-13B model in WMT22 translation tasks.}
    \label{fig:early-exit-with-diff-length}
\end{figure}

Our experiments show that existing gating networks are constrained by joint optimization, deciding whether to exit at each current layer leads to high computational costs in LLMs due to large vocabularies, lots of layers, and high hidden layer dimensions. Motivated by the natural early exit capability, we explore developing a gating network independent of joint optimization. Our analysis reveals that the optimal early exit layer decreases especially with shorter output sequences, as illustrated in Fig.~\ref{fig:early-exit-with-diff-length}. This suggests a potential reduction in the complexity of generation, consistent with the decreasing loss observed in~\cite{del2023skipdecode}. However, beyond a certain output length, the optimal early exit layer exhibited a slow and more unstable descent. To better understand this phenomenon, we conducted a detailed analysis from hidden state, sub-word, and part-of-speech perspectives.

\paragraph{Hidden State Similarity} We employ two metrics to measure the similarity of the hidden state and the output distribution from each layer, including the Jensen-Shannon divergence for distribution and the Cosine similarity for the hidden state. Our observations reveal that the similarity in the hidden state exhibits no significant trend even after joint optimization. In contrast, the similarity of output distributions becomes more different (detailed in~\ref{sec:layer-similarity}).

\paragraph{Sub-word} A notable trend emerged in the X-to-English translation direction: Approximately 12\% of tokens contribute to forming a complete word in all decoding tokens, and the first part of a word tends to exit in deeper layers, while the remaining part exits earlier, as illustrated in~\ref{sec:appendix-Trend-In-The-Optimal-Early-Exit-Layer-subword}. This means that the tokens at the beginning of a word are harder to predict while the rest are relatively easy, and this may also be the reason for the appearance of an unstable optimal early exit layer in Fig.\ref{fig:early-exit-with-diff-length}.

\paragraph{Part-of-speech} We observe significant variations in the optimal early exit layer across different UPOS labels (Tab.~\ref{tab:optimal-early-exit-layer-upos}). Similar to findings in Dola~\cite{chuang2023dola}, nouns and proper nouns closely tied to factual knowledge tend to have lower optimal early exit layers. In particular, symbols and adjectives also have lower optimal early exit layers, whereas determiner and punctuation are the opposite. This may be related to the frequency of vocabulary occurrences and their cross-linguistic consistency. For instance, symbols have lower exit layers maybe because of their consistent representation across languages.

\begin{table}[htp]
\small
\centering
\begin{tabular}{l|l|l|l|l|l}
\toprule
PROPN & NOUN & ADJ & SYM & DET & PUNCT\\
\midrule
21.23 & 19.79 & 19.39 & 19.11 & 25.65 & 26.77 \\
3.57\% & 10.86\% & 7.44\% & 0.12\% & 9.94\% & 8.17\% \\
\bottomrule
\end{tabular}
\caption{The optimal early exit layer the frequency of occurrence for parts UPOS Tags on WMT14 DE2EN (detailed in~\ref{sec:appendix-Trend-In-The-Optimal-Early-Exit-Layer-pos}).}
\label{tab:optimal-early-exit-layer-upos}
\end{table}

\subsection{Skip Connections Have High Consistency With Final Output}

\begin{figure}[htp]
    \centering
    \includegraphics[width=0.48\textwidth]{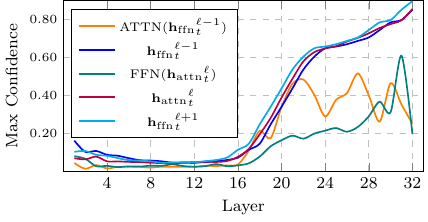}
    \caption{the Top-10 hypothesis launched by sub-layer in WMT22DE2EN.}
    \label{fig:top-hypothesis-and-confidence}
\vspace{-1em}
\end{figure}

We also attempt to extract the output from sub-layers, the FFN module and the ATTN module, inspired by relevant literature~\cite{geva2022transformer}. Our findings indicate that both the confidence score and the output token within the Top-10 hypotheses from the skip connection are stable. Conversely, the Top-10 hypotheses from the module consistently demonstrate substantial variations and notably smaller confidence scores compared to the skip connection, as depicted in Fig.~\ref{fig:top-hypothesis-and-confidence}. This can be approximated as the primary hypotheses being preserved in the skip connect, while the residual branch incrementally incorporates the most confident hypotheses into the primary branch layer by layer according to \cite{geva2022transformer}, and we omit the Softmax operation for simplicity:

The TOP output from the model has greater volatility, while its maximum output probability is lower~\ref{table:kv-example}
\begin{eqnarray}
& & {W}_{N} {h}^{\ell}_t = {W}_{N} \mathcal{F}({h}^{\ell-1}_t) + {W}_{N} {h}^{\ell-1}_t
\label{eq:top-hypothesis-in-sub-layer}
\end{eqnarray}

Further, we decode the same token and keep the decoding process exactly same with Fig.\ref{fig:saturation-event-in-llm} (c), and enumerated all Top-10 hypotheses from each layer, as shown in Table \ref{table:kv-example}. We find the consistent Top-1 hypothesis not only within the block output but also across the skip connection.

\section{Related Work}

\paragraph{Early Exit} Early exit is motivated by a hypothesis that some samples are easier to predict and require less computation costs~\cite{panda2016conditional, schwartz2020right, elbayad2019depth}. Implemented within the framework of joint optimization~\cite{teerapittayanon2016branchynet}, early exit is applied to both Bert~\cite{xin2020deebert, schwartz2020right} and Transformer~\cite{elbayad2019depth} models with a learnable depth estimator or estimate the required depths in advance~\cite{liu2021faster}. Although joint optimization leads to a multi-exit model (or called Once-for-All model~\cite{cai2019once}), two notable weaknesses of this framework are frequently addressed: 1) Joint optimization is challenging, as the loss from the shallow output layer may interfere with deeper output layer. Some researchers adapt dense connectivity~\cite{huang2017multi} or two-stage fine-tuning~\cite{xin2020deebert, xin2021berxit} to fix this problem. 2) An additional output layer may not output a good enough result. Facing this problem, ~\citet{liao2021global} proposes to combine both the past and future states from early and future layers to enhance current layer performance. ~\citet{zhou2020bert,gao2023f} determining exit layer by additional output layer continuously output the same result $n$ times. ~\citet{sun2021early} boost performance through ensemble methods involving multiple additional output layers. 

In the context of early exit research based on LLMs, ~\citet{kavehzadeh2023sorted} proposes to replace the standard supervised fine-tuning in joint optimization with sorted fine-tuning lead to a more flexible LLaMa model based on the SortedNet~\cite{valipour2023sortednet}. In the token-level early exit scenario, where the previous token exits earlier when generating the current token, a copied Key-Value caching must be used for self-attention for the current token (detail in \ref{sec:early-exit-back}). To avoid copying Key-Value caching and effectively use batched inference, ~\citet{del2023skipdecode} proposes to skip the early layer and exit only after the final layer. Additionally, ~\citet{bae2023fast} realizes a fast and robust early exit model by parallel decoding~\cite{leviathan2023fast} to avoid the KV caching problem. 

\paragraph{Saturation Events} Recently, a phenomenon called saturation events has provided strong support for early exit research, in which the final output prediction is consistently in the top-ranked prediction launched by the Feed-Forward Networks (FFN) in hidden layers and with increasing rank~\cite{geva2020transformer}. This implies that as the model uses more layers, the correct output that has been predicted by earlier layers gains more confidence, which motivated the CALM~\cite{schuster2022confident}. Moreover, it raises the possibility that the similarity between hidden states or distributions from each layer could help generate predictions~\cite{chuang2023dola}. The saturation events have further motivated us to research early exit. It is worth noting that we obtained the output after residual connection, not from the FFN update~\cite{geva2020transformer}, and we found a stronger saturation event in various tasks and transformer-based models. Additionally, we pay more attention to could this stronger saturation event is helpful for early exit.

\section{Conclusion}

The conventional early exit (EE) approach needs additional output layers and joint optimization hinders the application of EE in LLMs. In this paper, we explore the possibility of LLMs EE without additional output layers and joint optimization. Our findings indicate that EE is a natural capability within transformer-based models. While joint optimization does not give model EE capability, it must be employed to address challenges by improving the accuracy of locating the optimal EE layer through gating functions. Additionally, our study reveals patterns in EE behavior from a sub-word and part-of-speech perspective based on the LLaMA model and reveals the potential possibility for EE based on sub-layers.

\newpage
\section{Limitations}

While we do extensive experimentation with the Llama model, our research is currently constrained by the limitations of our available equipment, which has restricted us to a finite set of models. We look forward to expanding our experiment across a broader range of models in more resource scenarios, ensuring that our findings can be generalized to a wider array of environments.

\bibliography{custom}

\newpage

\appendix

\section{Detailed Experimental}
\label{sec:appendix}

\subsection{ALL Translation Result In WMT22 Dataset Based On The LLaMa2 Model}\label{sec:saturation-statistical-information}
We report the detailed LLaMa2 translation result on the WMT22 test set in Tab.~\ref{tab:saturation-in-all-task-model} based on the template shown as Tab.~\ref{tab:prompt-used-for-llm}. The COMET contains scores on ref.A \textbackslash ref.B respectively. We note Llama-2-Chat-7B as Llama7B, Llama-2-Chat-13B as Llama13B. We find that the model tends to have a lower average early exit layer at the 2EN translation direction.

\begin{table}[htp]
\small
\centering
\begin{tabular}{l|l|c|c}
\toprule
\multicolumn{2}{c|}{Model} & Llama7B & Llama13B \\
\midrule
\multicolumn{2}{c|}{Total layer} & \textbf{32} & \textbf{40} \\
\midrule
\multirow{3}{*}{DE$\rightarrow$EN} & COMET & 86.53\ \textbackslash\ 87.2 & 87.16\ \textbackslash\ 87.85 \\ & Avg & \textbf{22.44} & \textbf{22.62} \\ & Perc & 92.97 & 96.5 \\
\midrule
\multirow{3}{*}{DE$\rightarrow$FR} & COMET & 76.62 & 79.31 \\ & Avg & \textbf{24.67} & \textbf{26.74} \\ & Perc & 88.62 & 93.04 \\
\midrule
\multirow{3}{*}{EN$\rightarrow$DE} & COMET & 79.97\ \textbackslash\ 79.86 & 81.86\ \textbackslash\ 81.57 \\ & Avg & \textbf{24.47} & \textbf{26.96} \\ & Perc & 88.46 & 92.42 \\
\midrule
\multirow{3}{*}{EN$\rightarrow$JA} & COMET & 74.56 & 73.69 \\ & Avg & \textbf{23.34} & \textbf{25.39} \\ & Perc & 85.44 & 89.6 \\
\midrule
\multirow{3}{*}{EN$\rightarrow$RU} & COMET & 77.65 & 77.24 \\ & Avg & \textbf{24.35} & \textbf{26.81} \\ & Perc & 85.1 & 89.64 \\
\midrule
\multirow{3}{*}{EN$\rightarrow$ZH} & COMET & 77.47\ \textbackslash\ 78.48 & 77.11\ \textbackslash\ 78.15 \\ & Avg & \textbf{23.12} & \textbf{26.57} \\ & Perc & 85.16 & 87.64 \\
\midrule
\multirow{3}{*}{FR$\rightarrow$DE} & COMET & 77.31 & 80.57 \\ & Avg & \textbf{24.42} & \textbf{26.71} \\ & Perc & 89.36 & 92.75 \\
\midrule
\multirow{3}{*}{JA$\rightarrow$EN} & COMET & 80.31 & 81.49 \\ & Avg & \textbf{22.75} & \textbf{25.05} \\ & Perc & 91.65 & 93.95 \\
\midrule
\multirow{3}{*}{RU$\rightarrow$EN} & COMET & 85.81 & 86.51 \\ & Avg & \textbf{22.6} & \textbf{23.73} \\ & Perc & 92.06 & 95.19 \\
\midrule
\multirow{3}{*}{ZH$\rightarrow$EN} & COMET & 80.54\ \textbackslash\ 79.28 & 81.59\ \textbackslash\ 80.42 \\ & Avg & \textbf{22.8} & \textbf{24.45} \\ & Perc & 90.83 & 94.29 \\
\bottomrule
\end{tabular}
\caption{Detailed result in WMT22 translation task}
\label{tab:saturation-in-all-task-model}
\end{table}

\subsection{Results On Datasets Commonly Used For LLMs}\label{sec:commonly-used-dataset}

We performed the same experiment with Sec.~\ref{sec:Early-Exit-in-LLMs} on datasets commonly used for LLMs, including MBPP, GSM8K, and StrategyQA, shown in Tab.~\ref{tab:optimal-early-exit-layer-in-llama2-commonly-used-dataset}. We found that the optimal early exit layer showed a consistent trend across these common datasets, indicating that early exit capability is a very common phenomenon, and that it is widespread across a wide range of current tasks. In addition, we also find that as the length of the generated sequences gets longer, the average optimal early exit layer is lower, suggesting a greater potential for early exit in long text scenarios.

\begin{table}[htp]
\small
\centering
\begin{tabular}{l|c|c|c|c}
\toprule
\multirow{2}{*}{Model} & \multicolumn{3}{c|}{Early exit Layer} & \multirow{2}{*}{Length} \\ \cline{2-4}
& Max & Avg & Metrics \\
\midrule
\multicolumn{5}{c}{pass@1 on MBPP} \\
\midrule
7B & \textbf{32} & \textbf{20.19} & 0.299 & 128 \\
13B & \textbf{40} & \textbf{23.56} & 0.304 & 128 \\
\midrule
\multicolumn{5}{c}{Exact Match on GSM8K} \\
\midrule
7B & \textbf{32} & \textbf{20.56} & 0.1448 & 130.93 \\
13B & \textbf{40} & \textbf{24.45} & 0.2426 & 142.01 \\
\midrule
\multicolumn{5}{c}{Acc on StrategyQA} \\
\midrule
7B & \textbf{32} & \textbf{22.86} & 0.588 & 38.61 \\
13B & \textbf{40} & \textbf{26.11} & 0.609 & 38.61 \\
\bottomrule
\end{tabular}
\caption{The average optimal early exit layer and the average length of the generated sequences in various datasets.}
\label{tab:optimal-early-exit-layer-in-llama2-commonly-used-dataset}
\end{table}

\subsection{Template For LLaMa2 Inference}\label{sec:template-for-prompt}

To verify our decoding experiment we try the ALMA style prompt and keep all other settings, as shown in Tab.~\ref{tab:prompt-used-for-llm}. We got the same result with the LLaMA-2-7B (zero-shot) reported in paper~\cite{xu2023paradigm}. In our experiment, we only changed the demonstration in our prompt and did not change other hyperparameters.

\begin{table*}[htp]
\small
    \centering
    \begin{tabular}{c|p{0.7\textwidth}}  
    \toprule
    \textbf{Tasks} & \textbf{System Prompt} \\
    
    \midrule
    \multirow{7}{*}{Translation} & \#\#\# Instruction:\\ & Translate \texttt{src} to \texttt{tgt}:\\ & \\ & \#\#\# Input:\\ & \texttt{real input}\\ & \\ & \#\#\# Response: \\
    \midrule
    \multirow{7}{*}{Summarization} & \#\#\# Instruction:\\ & Summarize the following article to a sentence:\\ & \\ & \#\#\# Input:\\ & \texttt{real input}\\ & \\ & \#\#\# Response: \\
    \midrule
    \multirow{8}{*}{Question Answering} & \#\#\# Instruction: I will provide a context and a question to you. You need to answer me the question based on the context.\\ & \\ & \#\#\# Context: \texttt{The context}\\ & \\ & \#\#\# Question: \texttt{Question}\\ & \\ & \#\#\# Answer: \\
    \bottomrule
    \end{tabular}
    \caption{Prompts for generation task. For translation tasks, \texttt{src} and \texttt{tgt} is select from \{English, Chinese, German, Russian, French, Japanese\}, and the \texttt{real input} is the sentence to be translated}
    \label{tab:prompt-used-for-llm}
\end{table*}

\begin{figure*}[htp]
    \centering
    \includegraphics[width=1.0\textwidth]{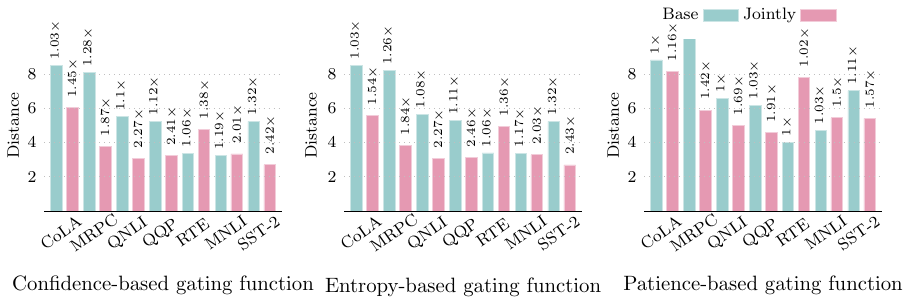}
    \caption{Roberta model.}
    \label{fig:direct-exit-by-previous-function-roberta}
\end{figure*}

\section{Gating Functions On The Backbone Model}

\subsection{BERT And RoBERTa Model}\label{sec:appendix-gating-function directly-on-bert-like-model}

we have shown the distance and speed-up between the optimal early exit layer and the exit layer in detail, the threshold chosen to constrain the performance of early exit not less than 98\% of the original model, is illustrated in Fig \ref{fig:direct-exit-by-previous-function-roberta}. Across almost all tasks, joint optimization significantly enhances the accuracy of the gating function by improving the similarity of distribution output from each layer. While the gap is relatively small in the patience-based gating function, this can be attributed to the gating function exiting only at $n$-th continuous layer, which yields the same result, leading to a deeper exit layer and a more accurate result normally. Additionally, a noteworthy observation is the decline in the average optimal early exit layer across most tasks after joint optimization, signifying an improvement in the upper bound of early exit.

\begin{table*}[htp]
\small
\centering
\begin{tabular}{l|c|c|c|c}
\toprule
Model & \multicolumn{2}{c|}{RoBERTa(\textbf{12})} & \multicolumn{2}{c}{BERT(\textbf{12})} \\ 
\midrule
Metric & Avg & Perc & Avg & Perc \\
\midrule
CoLA & 3.17 -> \textbf{2.26} & 81.11\% -> 99.90\% & 3.11 -> \textbf{2.40} & 93.10\% -> 99.90\% \\
MRPC & 1.32 -> \textbf{2.94} & 100.0\% -> 99.88\% & 2.88 -> \textbf{2.21} & 93.68\% -> 99.83\% \\
QNLI & 5.44 -> \textbf{2.27} & 92.97\% -> 99.96\% & 2.87 -> \textbf{1.93} & 99.65\% -> 99.80\% \\
QQP & 5.50 -> \textbf{1.83} & 99.82\% -> 99.99\% & 2.04 -> \textbf{1.62} & 100.0\% -> 99.95\% \\
RTE & 8.00 -> \textbf{3.92} & 77.26\% -> 99.64\% & 3.60 -> \textbf{1.62} & 96.39\% -> 99.28\% \\
MNLI & 7.01 -> \textbf{2.75} & 98.68\% -> 99.96\% & 6.46 -> \textbf{2.35} & 93.40\% -> 99.84\% \\
SST-2 & 3.83 -> \textbf{2.34} & 100.0\% -> 99.77\% & 4.67 -> \textbf{1.58} & 99.77\% -> 100.0\% \\
\bottomrule
\end{tabular}
 \caption{The optimal early exit layer in roberta-base and BERT-base-uncased after joint optimization.}
\label{tab:saturation-in-bert}
\end{table*}

\subsection{Transformer And LLaMa2 Model}\label{sec:appendix-gating-function directly-on-transformer-and-llm}

We performed the same experiment on the Transformer and LLaMa2 model described as \ref{sec:appendix-gating-function directly-on-bert-like-model}. We find that the transformer model with the joint optimization can select an exit layer that is very close to the optimal early exit layer, and lead to a significant speed up to the inference process. For the llama model, the joint optimization still improves the effectiveness of the gating function, but the exit layer output by the gating function mentioned above is still far from the optimal early exit layer, which indicates that although the joint optimization helps to improve the performance of the gating function, it still cannot reach the optimal early exit layers.

\begin{figure*}[htp]
    \centering
    \includegraphics[width=1.0\textwidth]{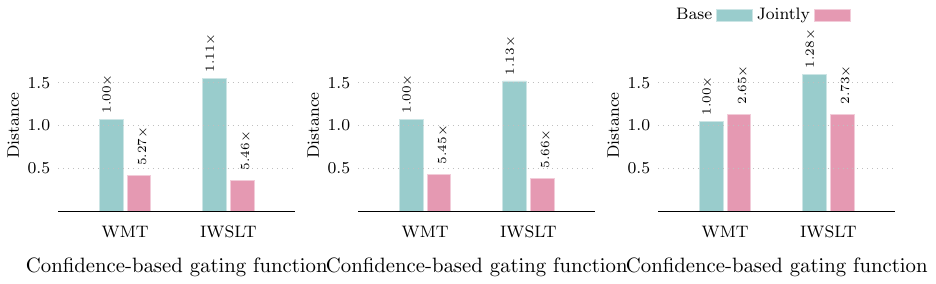}
    \caption{Transformer-base model follow DLCL~\cite{wang2019learning} on the WMT14-EN2DE and DAT~\cite{elbayad2019depth} on the IWSLT14-DE2EN.}
    \label{fig:direct-exit-by-previous-function-Transformer-model}
\end{figure*}

\begin{figure*}[htp]
    \centering
    \includegraphics[width=1.0\textwidth]{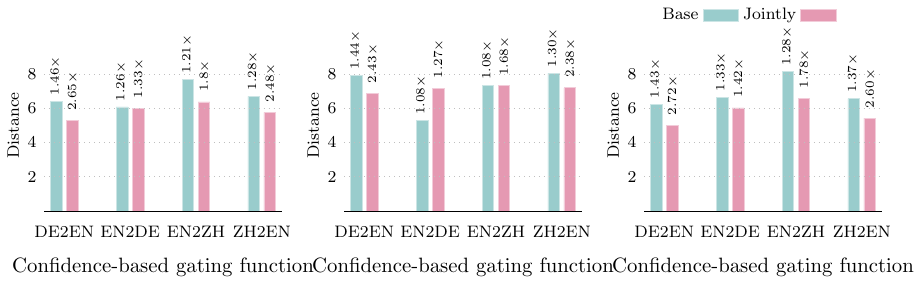}
    \caption{LLaMa2 model on the WMT22 test set.}
    \label{fig:direct-exit-by-previous-function-llama-model}
\end{figure*}

\section{Similarity In Each Layer}\label{sec:layer-similarity}

\begin{figure}[htp]
    \centering
    \includegraphics[scale=0.35]{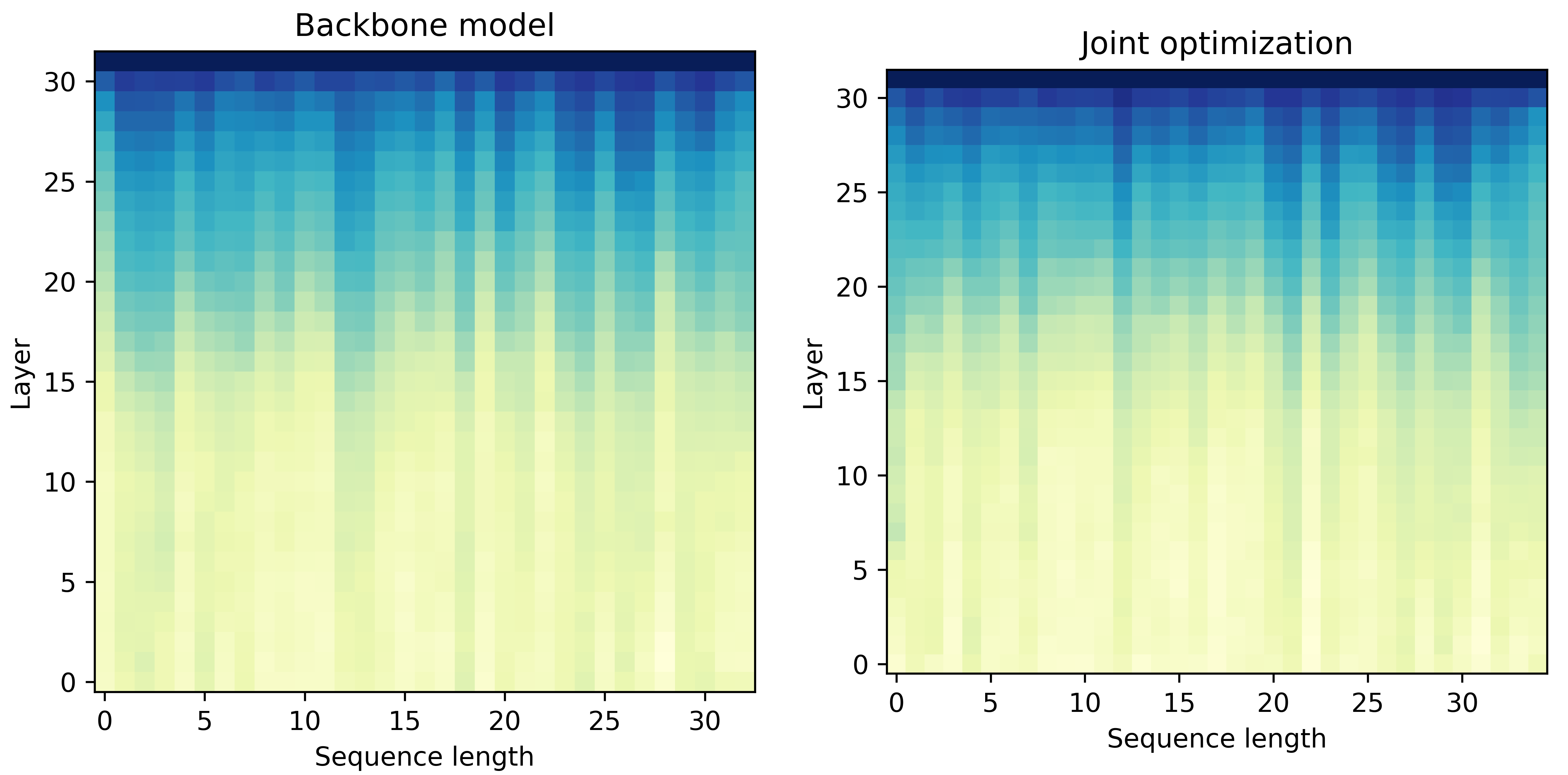}
    \includegraphics[scale=0.35]{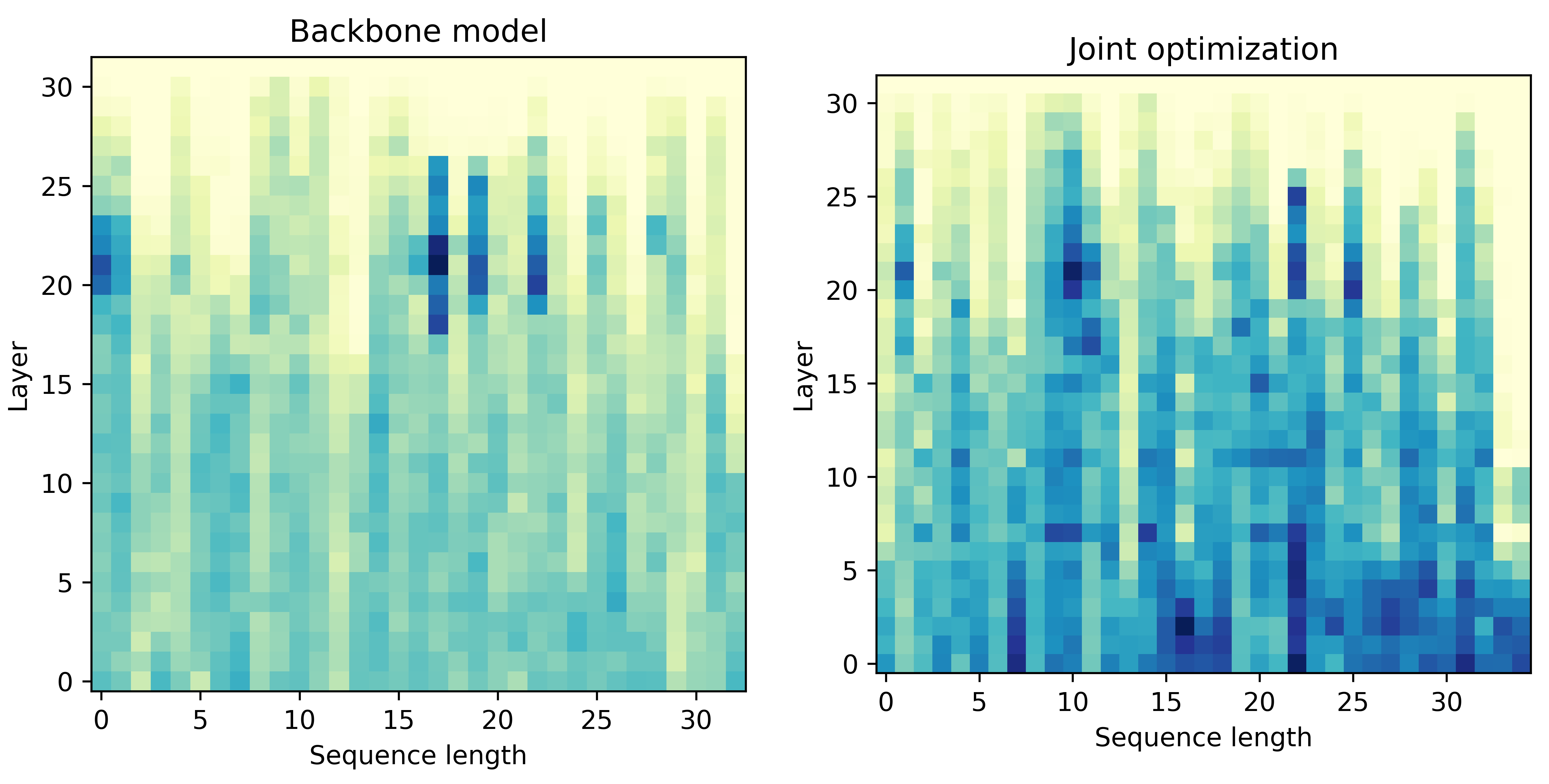}
    \caption{Cosine similarity of hidden state (up) and Js divergence of distribution (down) between each layer on the same sentence. The darker the color the greater the value.}
    \label{fig:similarity-of-distribution}
\end{figure}

We find no significant difference in cos similarity between the hidden state from each layer, as shown in Fig.~\ref{fig:similarity-of-distribution}. Whereas in the distribution, the model with joint optimization tends to have higher similarity in the deeper layers, while more obvious difference in the shallow layers, which is in line with our observation in Sec.~\ref{sec:Joint-Helps-Gating-by-Boosting-Output-Distribution-Similarity}. In the actual decoding process we also tried to use the cosine similarity from different hidden states to determine whether we could output at the current layer, but the results were unacceptable. Even with the carefully tuned thresholds in our experiments, we could not determine whether the current layer would lead to a good output.

\section{Property Of Optimal Early Exit Layer}

\subsection{Sub-word}\label{sec:appendix-Trend-In-The-Optimal-Early-Exit-Layer-subword}

In the translation model, complete words are segmented into sub-words, to explore whether the positions of sub-words affect the number of optimal early exit layers, we labeled a sub-word at the beginning of a word as the word's prefix, with the rest of the word as suffixes. We found that in different languages the prefix of the word tends to require more layers to prediction, indicating that generating the beginning of a word is more difficult, and as the initial part is determined, the rest of the word may be easier to generate.

\begin{figure*}[htp]
    \centering
    \includegraphics[width=1.0\textwidth]{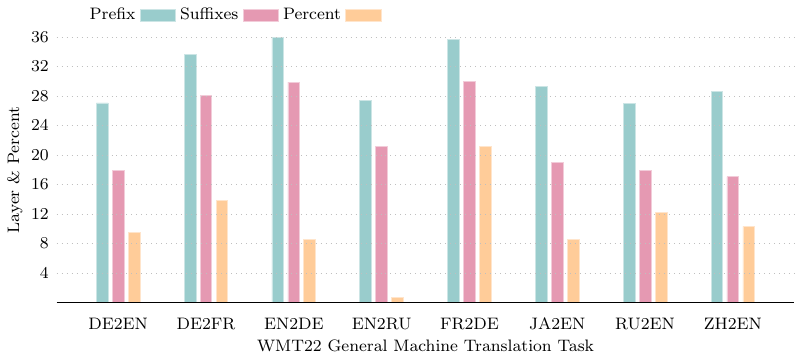}
    \caption{Translation result of LLaMa2 model on the WMT22 test set. We use prefix to represent the initial segment of a word and suffixes for the remaining part. Percent indicates the percentage of this phenomenon occurrence in all tokens.}
    \label{fig:prefix-llama-model}
\end{figure*}

\subsection{Part of speech}\label{sec:appendix-Trend-In-The-Optimal-Early-Exit-Layer-pos}

We also used the Stanford CoreNLP toolkit\footnotemark[2]\footnotetext[2]{\url{https://github.com/stanfordnlp/CoreNLP}} to compute the part of speech of different words, we express the average optimal early exit layer of the specific part of speech by the average optimal early exit layer of all tokens which have the same part of speech. We found that not the words with higher frequency lead lower average early exit layer. Among the words with higher frequency, nouns, and adjectives have a lower optimal average early exit layer. On the contrary, punctuation with 0.12\% frequency has the lowest average early exit layer among all parts of speech.

\begin{table}[htp]
\small
\centering
\begin{tabular}{l|l|l|l|l|l}
\toprule
DET & NOUN & VERB & ADJ & ADP & NUM\\
\midrule
25.65 & 19.79 & 23.60 & 19.39 & 22.21 & 22.59 \\
9.94\% & 11.04\% & 10.86\% & 7.44\% & 9.70\% & 2.23\% \\
\midrule
PUNCT & PRON & AUX & CCONJ & ADV & PART\\
\midrule
26.77 & 20.60 & 23.14 & 21.25 & 20.69 & 20.56 \\
8.17\% & 8.17\% & 7.94\% & 4.98\% & 6.92\% & 2.90\% \\
\midrule
SCONJ & PROPN & INTJ & SYM & X & -\\
\midrule
21.68 & 21.24 & 21.74 & 19.11 & 22.17 & - \\
2.54\% & 3.57\% & 0.36\% & 0.12\% & 0.01\% & - \\
\bottomrule
\end{tabular}
\caption{The optimal early exit layer and the frequency of occurrence for all UPOS Tags on WMT14 DE2EN.}
\label{tab:optimal-early-exit-layer-all-upos}
\end{table}

\subsection{Top-10 Hypotheses From Sub-layer And Module}\label{sec:appendix-Top-Hypothesis-From-Sub-layer-And-Module}

We decode the same token and keep the decoding process exactly the same with Figure \ref{fig:saturation-event-in-llm} (c), and enumerated all Top-10 hypotheses from each layer which consistently produces the same Top-1 output as the final layer, as shown in Table \ref{table:kv-example}. We find the consistent Top-1 hypothesis from the $20$-th layer to the $40$-th layer, not only within the block output but also across the skip connection. However, the final output \textit{\_fields} rarely surfaced in the Top-10 hypotheses from the module output, but it appears occasionally and improves the rank of final output in the Top-10 hypotheses like the \textit{\_fields} in $\text{FFN}(\mathbf{h_{\text{attn}}}^{23}_t)$.

\begin{table*}[ht]
\small
\centering
\begin{tabular}{ll}
\toprule
\textbf{Exit position} & \textbf{Top-10 hypotheses} \\
\midrule
$\mathbf{h_{\text{ffn}}}^{40}_t$ & \_fields, \_areas, \_discipl, \_domains, \_inter, fields, \_field, \_subjects, \_major, \_indust \\
$\mathbf{h_{\text{attn}}}^{40}_t$ & \_fields, \_areas, \_discipl, \_inter, \_field, fields, \_major, \_se, \_domains, \_subjects \\
$\text{FFN}(\mathbf{h_{\text{attn}}}^{40}_t)$ & <s>, \_edific, textt, \_departamento, metros, \_religios, \_communes, \textendash, \_interfaces, Portail \\
$\mathbf{h_{\text{ffn}}}^{39}_t$ & \_fields, \_areas, \_discipl, \_inter, fields, \_field, \_domains, \_major, \_subjects, \_se \\
$\text{ATTN}(\mathbf{h_{\text{ffn}}}^{39}_t)$ & <0x0A>, ,, \_, -, ., \_and, \_(, \_the, \_in, \_C \\
\midrule
$\mathbf{h_{\text{ffn}}}^{39}_t$ & \_fields, \_areas, \_discipl, \_inter, fields, \_field, \_domains, \_major, \_subjects, \_se \\
$\mathbf{h_{\text{attn}}}^{39}_t$ & \_fields, \_areas, \_discipl, \_field, fields, \_domains, \_inter, \_subjects, \_major, \_maj \\
$\text{FFN}(\mathbf{h_{\text{attn}}}^{39}_t)$ & \_, \_A, \_(, \_C, <0x0A>, \_R, \_just, \_in, \_g, \_G \\
$\mathbf{h_{\text{ffn}}}^{38}_t$ & \_fields, \_areas, \_discipl, \_field, fields, \_domains, \_inter, \_subjects, \_major, \_maj \\
$\text{ATTN}(\mathbf{h_{\text{ffn}}}^{38}_t)$ & \_…, \_..., \_., \_, \_covering, \_cover, ,, \_and, \_coverage, \_.. \\
\midrule
$\vdots$ & $\vdots$ \\
\midrule
$\mathbf{h_{\text{ffn}}}^{23}_t$ & \_fields, \_areas, \_field, \_discipl, \_categories, fields, \_dici, \_domains, \_subjects, \_topics \\
$\mathbf{h_{\text{attn}}}^{23}_t$ & \_fields, \_areas, \_field, \_categories, \_discipl, \_domains, fields, Fields, \_topics, \_dici \\
$\text{FFN}(\mathbf{h_{\text{attn}}}^{23}_t)$ & \_subject, aban, \_rising, \_fields, engo, enten, gew, chten, \_nich, \_branches \\
$\mathbf{h_{\text{ffn}}}^{22}_t$ & \_areas, \_fields, \_discipl, \_categories, \_domains, \_area, \_topics, \_aspects, \_dici, \_major \\
$\text{ATTN}(\mathbf{h_{\text{ffn}}}^{22}_t)$ & \_field, \_fields, \_Field, Field, fields, field, Fields, \_research, \_none, \_campo \\
\midrule
$\vdots$ & $\vdots$ \\
\midrule
$\mathbf{h_{\text{ffn}}}^{20}_t$ & \_fields, \_areas, \_aspects, \_major, \_categories, \_topics, \_discipl, \_types, \_subjects, \_area \\
$\mathbf{h_{\text{attn}}}^{20}_t$ & \_areas, \_fields, \_aspects, \_topics, \_subjects, \_categories, \_major, \_discipl, \_types, \_area \\
$\text{FFN}(\mathbf{h_{\text{attn}}}^{20}_t)$ & yl, \_natural, \_str, \_kind, \_pure, \_proven, \_un, \_extreme, \_underlying, \_flav \\
$\mathbf{h_{\text{ffn}}}^{19}_t$ & \_areas, \_fields, \_aspects, \_topics, \_categories, \_subjects, \_types, \_major, \_discipl, \_area \\
$\text{ATTN}(\mathbf{h_{\text{ffn}}}^{19}_t)$ & \_territ, \_fields, \_field, cipl, \_sector, \_discipline, \_territory, sci, \_domains, \_indust \\
\bottomrule
\end{tabular}
\caption{Top-10 hypotheses from block output $\mathbf{h_{\text{ffn}}}^{\ell+1}_t$, ffn module $\text{FFN}(\mathbf{h_{\text{attn}}}^{\ell}_t)$, attn module $\text{ATTN}(\mathbf{h_{\text{ffn}}}^{\ell-1}_t)$ and skip connect $\mathbf{h_{\text{attn}}}^{\ell}_t, \mathbf{h_{\text{ffn}}}^{\ell-1}_t$ based on the final output layer.}
\label{table:kv-example}
\end{table*}

\section{Copy KV Cache In longer sentence}

For applying early exit in long sequences generation scenario, we find recomputing KV works well on the model without joint optimization. However recomputing KV seems can not work at all on the model with joint optimization.

\begin{table*}[ht]
\centering
\footnotesize
\begin{tabular}{lp{4.5cm}p{9.2cm}}
\toprule
& \textbf{SRC} & \textbf{Denn falls tatsächlich etwas passieren sollte wie ein Brand, Einbruch, Erdbeben, Alieninvasion etc. wäre es tatsächlich zu viel Verantwortung für K1, sich um K2 zu kümmern.} \\
\midrule
& hypothesis & optimal early exit layer \\ \midrule
7B & If something were to happen like a fire, burglary, earthquake, or alien invasion, it would be too much responsibility for K1 to take care of K2.</s> & \texttt{[31, 31, 31, 31, 31, 31, 31, 31, 31, 31, 31, 31, 31, 31, 31, 31, 31, 31, 31, 31, 31, 31, 31, 31, 31, 31, 31, 31, 31, 31, 31, 31, 31, 31, 31, 31, 31, 31, 31]} \\ 
7B-r & If something were to happen like a fire, theft, earthquake, or alien invasion, it would be too much for K1\&elo to take care of K2 inoculation.i</s> & \texttt{[31, 31, 31, 31, 5, 16, 26, 16, 16, 31, 10, 22, 16, 15, 19, 31, 18, 24, 0, 16, 30, 17, 13, 22, 15, 1, 30, 19, 17, 30, 30, 30, 29, 0, 24, 24, 16, 30, 31, 29, 29, 31, 27]} \\ 
\midrule
7B-j &  If something were to happen like a fire, break-in, earthquake, or alien invasion, it would be too much responsibility for K1 to take care of K2.</s> & \texttt{[31, 31, 31, 31, 31, 31, 31, 31, 31, 31, 31, 31, 31, 31, 31, 31, 31, 31, 31, 31, 31, 31, 31, 31, 31, 31, 31, 31, 31, 31, 31, 31, 31, 31, 31, 31, 31, 31, 31]} \\ 
7B-j-r & If something were to happen like a fire, a (catastrophic) alien invasion, (a)n (alien) invasion, (a)nd (al (alien), (al (al), (al (al), (al (al), (al (al), (al (al), (al (al), (al (al), (al (al), (al (al), (al (al).</s> & \texttt{[31, 31, 31, 31, 5, 16, 7, 10, 7, 7, 31, 31, 8, 2, 2, 8, 31, 0, 7, 7, 15, 23, 5, 30, 30, 30, 13, 7, 15, 7, 6, 6, 4, 24, 7, 8, 31, 12, 15, 29, 30, 12, 31, 9, 31, 6, 12, 17, 9, 7, 6, 12, 7, 9, 7, 6, 11, 7, 9, 7, 6, 11, 7, 9, 7, 6, 11, 7, 9, 7, 6, 11, 7, 9, 7, 6, 11, 7, 9, 7, 6, 11, 7, 9, 7, 6, 11, 7, 9, 27, 7]} \\ 
\bottomrule
\end{tabular}
\caption{Token-level early exit result and exit layer of Llama-2-Chat-7B(7B) with(-J) and without joint optimization on WMT22-DE2EN test set, under the constraint that early exit only after the 4-th token is generated. For copy KV operation, we represent recompute as -r.}
\label{table:kv-example-in-long-sentence}
\end{table*}

\end{document}